\documentclass[11pt,a4paper]{article}
\usepackage[hyperref]{emnlp-ijcnlp-2019}
\usepackage{times}
\usepackage{latexsym}

\usepackage[pdftex]{graphicx}

\usepackage{algorithmic}
\usepackage{algorithm}

\usepackage{url}
\usepackage{amsmath}
\usepackage{amsfonts}

\DeclareMathOperator*{\argmax}{arg\,max}

\aclfinalcopy 

\newcommand{\customfootnotetext}[2]{{
  \renewcommand{\thefootnote}{#1}
  \footnotetext[0]{#2}}}

\title{On NMT Search Errors and Model Errors: Cat Got Your Tongue?}

\author{Felix Stahlberg$^*$ \and Bill Byrne \\
  University of Cambridge \\
  Department of Engineering \\
  Trumpington St, Cambridge CB2 1PZ, UK \\
  {\tt \{fs439,wjb31\}@cam.ac.uk}}

\date{}

\begin{document}
\maketitle
\begin{abstract}
  We report on search errors and model errors in neural machine translation (NMT). We present an exact inference procedure for neural sequence models based on a combination of beam search and depth-first search. We use our exact search to find the global best model scores under a Transformer base model for the entire WMT15 English-German test set. Surprisingly, beam search fails to find these global best model scores in most cases, even with a very large beam size of 100. For more than 50\% of the sentences, the model in fact assigns its global best score to the empty translation, revealing a massive failure of neural models in properly accounting for adequacy. We show by constraining search with a minimum translation length that at the root of the problem of empty translations lies an inherent bias towards shorter translations. We conclude that vanilla NMT in its current form requires just the right amount of beam search errors, which, from a modelling perspective, is a highly unsatisfactory conclusion indeed, as the model often prefers an empty translation.
\end{abstract}

\section{Introduction}

\customfootnotetext{*}{Now at Google.}

Neural machine translation~\citep[NMT]{nnjm-recurrent,nmt-sutskever,nmt-bahdanau} assigns the probability $P(\mathbf{y}|\mathbf{x})$ of a translation $\mathbf{y}=y_1^J\in \mathcal{T}^J$ of length $J$ over the target language vocabulary $\mathcal{T}$ for a source sentence $\mathbf{x}\in \mathcal{S}^I$ of length $I$ over the source language vocabulary $\mathcal{S}$ via a left-to-right factorization using the chain rule:
\begin{equation}
\log P(\mathbf{y}|\mathbf{x}) = \sum_{j=1}^J \log P(y_j|y_1^{j-1},\mathbf{x}).
\label{eq:nmt}
\end{equation}
The task of finding the most likely translation $\hat{\mathbf{y}}\in\mathcal{T}^*$ for a given source sentence $\mathbf{x}$ is known as the {\em decoding} or {\em inference} problem:
\begin{equation}
\hat{\mathbf{y}}=\argmax_{\mathbf{y}\in \mathcal{T}^*} P(\mathbf{y}|\mathbf{x}).
\label{eq:nmt-inference}
\end{equation}
The NMT search space is vast as it grows exponentially with the sequence length. For example, for a common vocabulary size of $|\mathcal{T}|=32,000$, there are already more possible translations with 20 words or less than atoms in the observable universe ($32,000^{20}\gg 10^{82}$). Thus, complete enumeration of the search space is impossible. The size of the NMT search space is perhaps the main reason why -- besides some preliminary studies~\citep{nmt-search-vs-model-errors,sgnmt2,conf-ana-uncertainty} -- analyzing search errors in NMT has received only limited attention. To the best of our knowledge, none of the previous studies were able to quantify the number of search errors in unconstrained NMT due to the lack of an exact inference scheme that -- although too slow for practical MT -- guarantees to find the global best model score for analysis purposes.

\begin{algorithm*}[t!]
\caption{BeamSearch$(\mathbf{x},n\in \mathbb{N}_+)$}
\label{alg:nmt-search-beam}
\begin{algorithmic}[1]
\REQUIRE{$\mathbf{x}$: Source sentence, $n$: Beam size}
\STATE{$\mathcal{H}_{cur}\gets \{(\epsilon, 0.0)\}$} \COMMENT{Initialize with empty translation prefix and zero score}
\REPEAT
  \STATE{$\mathcal{H}_{next}\gets \emptyset$}
  \FORALL{$(\mathbf{y},p)\in \mathcal{H}_{cur}$}
    \IF{$y_{|\mathbf{y}|}=\mathtt{</s>}$} 
      \STATE{$\mathcal{H}_{next}\gets \mathcal{H}_{next} \cup \{(\mathbf{y},p)\}$}
      \COMMENT{Hypotheses ending with $\mathtt{</s>}$ are not expanded}
    \ELSE
      \STATE{$\mathcal{H}_{next}\gets \mathcal{H}_{next} \cup \bigcup_{w\in \mathcal{T}} (\mathbf{y}\cdot w, p + \log P(w|\mathbf{x},\mathbf{y}))$}
      \COMMENT{Add all possible continuations}
    \ENDIF
  \ENDFOR
  \STATE{$\mathcal{H}_{cur}\gets \{(\mathbf{y},p)\in \mathcal{H}_{next} : |\{(\mathbf{y}',p')\in \mathcal{H}_{next} : p' > p\}| < n\}$}
  \COMMENT{Select $n$-best}
  \STATE{$(\tilde{\mathbf{y}},\tilde{p}) \gets \argmax_{(\mathbf{y},p)\in \mathcal{H}_{cur}} p$}
\UNTIL{$\tilde{y}_{|\tilde{\mathbf{y}}|} = \mathtt{</s>}$}
\RETURN{$\tilde{\mathbf{y}}$}
\end{algorithmic}
\end{algorithm*}

\begin{algorithm}[t!]
\caption{DFS$(\mathbf{x},\mathbf{y},p\in\mathbb{R},\gamma\in \mathbb{R})$}
\label{alg:nmt-search-dfs}
\begin{algorithmic}[1]
\REQUIRE{$\mathbf{x}$: Source sentence \\
\hspace{1.5em}$\mathbf{y}$: Translation prefix (default: $\epsilon$) \\
\hspace{1.5em}$p$: $\log P(\mathbf{y}|\mathbf{x})$ (default: $0.0$) \\
\hspace{1.5em}$\gamma$: Lower bound}
\IF{$y_{|\mathbf{y}|}=\mathtt{</s>}$} 
  \RETURN{$(\mathbf{y},p)$} \COMMENT{Trigger $\gamma$ update}
\ENDIF
\STATE{$\tilde{\mathbf{y}}\gets \perp$} \COMMENT{Initialize $\tilde{\mathbf{y}}$ with dummy value}
\FORALL{$w\in\mathcal{T}$}
  \STATE{$p'\gets p + \log P(w|\mathbf{x},\mathbf{y})$}
  \IF{$p'\geq \gamma$}
    \STATE{$(\mathbf{y}',\gamma')\gets \text{DFS}(\mathbf{x}, \mathbf{y}\cdot w, p', \gamma)$}
    \IF{$\gamma'> \gamma$}
        \STATE{$(\tilde{\mathbf{y}},\gamma)\gets (\mathbf{y}',\gamma')$}
    \ENDIF
  \ENDIF
\ENDFOR
\RETURN{$(\tilde{\mathbf{y}},\gamma)$}
\end{algorithmic}
\end{algorithm}

In this work we propose such an exact decoding algorithm for NMT that exploits the monotonicity of NMT scores: Since the conditional log-probabilities in Eq.~\ref{eq:nmt} are always negative, partial hypotheses can be safely discarded once their score drops below the log-probability of any {\em complete} hypothesis. Using our exact inference scheme we show that beam search does not find the global best model score for more than half of the sentences. However, these {\em search} errors, paradoxically, often prevent the decoder from suffering from a frequent but very serious {\em model} error in NMT, namely that the empty hypothesis often gets the global best model score. Our findings suggest a reassessment of the amount of model and search errors in NMT, and we hope that they will spark new efforts in improving NMT modeling capabilities, especially in terms of adequacy.

\section{Exact Inference for Neural Models}

Decoding in NMT (Eq.~\ref{eq:nmt-inference}) is usually tackled with beam search, which is a time-synchronous approximate search algorithm that builds up hypotheses from left to right. A formal algorithm description is given in Alg.~\ref{alg:nmt-search-beam}. Beam search maintains a set of active hypotheses $\mathcal{H}_{cur}$. In each iteration, all hypotheses in $\mathcal{H}_{cur}$ that do not end with the end-of-sentence symbol $\mathtt{</s>}$ are expanded and collected in $\mathcal{H}_{next}$. The best $n$ items in $\mathcal{H}_{next}$ constitute the set of active hypotheses $\mathcal{H}_{cur}$ in the next iteration (line 11 in Alg.~\ref{alg:nmt-search-beam}), where $n$ is the beam size. The algorithm terminates when the best hypothesis in $\mathcal{H}_{cur}$ ends with the end-of-sentence symbol $\mathtt{</s>}$. Hypotheses are called {\em complete} if they end with $\mathtt{</s>}$ and {\em partial} if they do not.

Beam search is the ubiquitous decoding algorithm for NMT, but it is prone to search errors as the number of active hypotheses is limited by $n$. In particular, beam search never compares partial hypotheses of different lengths with each other. As we will see in later sections, this is one of the main sources of search errors. However, in many cases, the model score found by beam search is a reasonable approximation to the global best model score. Let $\gamma$ be the model score found by beam search ($\tilde{p}$ in line 12, Alg.~\ref{alg:nmt-search-beam}), which is a lower bound on the global best model score: $\gamma \leq \log P(\hat{\mathbf{y}}|\mathbf{x})$. Furthermore, since the conditionals $\log P(y_j|y_1^{j-1},\mathbf{x})$ in Eq.~\ref{eq:nmt} are log-probabilities and thus non-positive, expanding a partial hypothesis is guaranteed to result in a lower model score, i.e.:\footnote{Equality in Eq.~\ref{eq:score-monotone} is impossible since probabilities are modeled by the neural model via a softmax function which never predicts a probability of {\em exactly} 1.}
\begin{equation}
\forall j\in[2,J]: \log P(y_1^{j-1}|\mathbf{x}) > \log P(y_1^j|\mathbf{x}).
\label{eq:score-monotone}
\end{equation}
Consequently, when we are interested in the global best hypothesis $\hat{\mathbf{y}}$, we only need to consider partial hypotheses with scores greater than $\gamma$. In our exact decoding scheme we traverse the NMT search space in a depth-first order, but cut off branches along which the accumulated model score falls below $\gamma$. During depth-first search (DFS), we update $\gamma$ when we find a better complete hypothesis. Alg.~\ref{alg:nmt-search-dfs} specifies the DFS algorithm formally. An important detail is that elements in $\mathcal{T}$ are ordered such that the loop in line 5 considers the $\mathtt{</s>}$ token first. This often updates $\gamma$ early on and leads to better pruning in subsequent recursive calls.\footnote{Note that the order in which the for-loop in line 5 of Alg.~\ref{alg:nmt-search-dfs} iterates over $\mathcal{T}$ may be important for efficiency but does not affect the correctness of the algorithm.}

\paragraph{Exact inference under length constraints}

Our admissible pruning criterion based on $\gamma$ relies on the fact that the model score of a (partial) hypothesis is always lower than the score of any of its translation prefixes. While this monotonicity condition is true for vanilla NMT (Eq.~\ref{eq:score-monotone}), it does not hold for methods like length normalization~\citep{sys-montreal-wmt15,nn-length-norm,production-gnmt} or word rewards~\citep{hybrid-nmt-with-smt-features}: Length normalization gives an advantage to longer hypotheses by dividing the score by the sentence length, while a word reward directly violates monotonicity as it rewards each word with a positive value.  In Sec.~\ref{sec:length-constraints} we show how our exact search can be extended to handle arbitrary length models~\citep{nmt-correcting-lengthbias,nmt-optimal-beam,nmt-stopping-criteria} by introducing length dependent lower bounds $\gamma_k$ and report initial findings on exact search under length normalization. However, despite being of practical use, methods like length normalization and word penalties are rather heuristic as they do not have any justification from a probabilistic perspective. They also do not generalize well as (without retuning) they often work only for a specific beam size. It would be much more desirable to fix the length bias in the NMT model itself. 

\section{Results without Length Constraints}

We conduct all our experiments in this section on the entire English-German WMT \texttt{news-test2015} test set (2,169 sentences)  with a Transformer base~\citep{nmt-transformer} model trained with Tensor2Tensor~\citep{nmt-tool-t2t} on parallel WMT18 data excluding ParaCrawl. Our pre-processing is as described by~\citet{ucam-wmt18} and includes joint subword segmentation using byte pair encoding~\citep{nmt-bpe} with 32K merges. We report cased BLEU scores.\footnote{Comparable with \url{http://matrix.statmt.org/}} An open-source implementation of our exact inference scheme is available in the SGNMT decoder~\citep{sgnmt1,sgnmt2}.\footnote{\url{http://ucam-smt.github.io/sgnmt/html/}, \texttt{simpledfs} decoding strategy.}

\begin{table}
\centering
\small
\begin{tabular}{|l|r@{\hspace{0.9em}}r@{\hspace{0.9em}}r@{\hspace{0.9em}}r|}\hline
\textbf{Search} & \textbf{BLEU} & \textbf{Ratio} & \textbf{\#Search errors} & \textbf{\#Empty} \\ \hline
Greedy & 29.3 & 1.02 & 73.6\% & 0.0\% \\
Beam-10 & 30.3 & 1.00 & 57.7\% & 0.0\% \\
Exact & 2.1 & 0.06 & 0.0\% & 51.8\% \\
    \hline
\end{tabular}
\caption{NMT with exact inference. In the absence of search errors, NMT often prefers the empty translation, causing a dramatic drop in length ratio and BLEU.}\label{tab:exact-search}
\end{table}

\begin{figure}[t!]
\centering
\small
\includegraphics[scale=0.85]{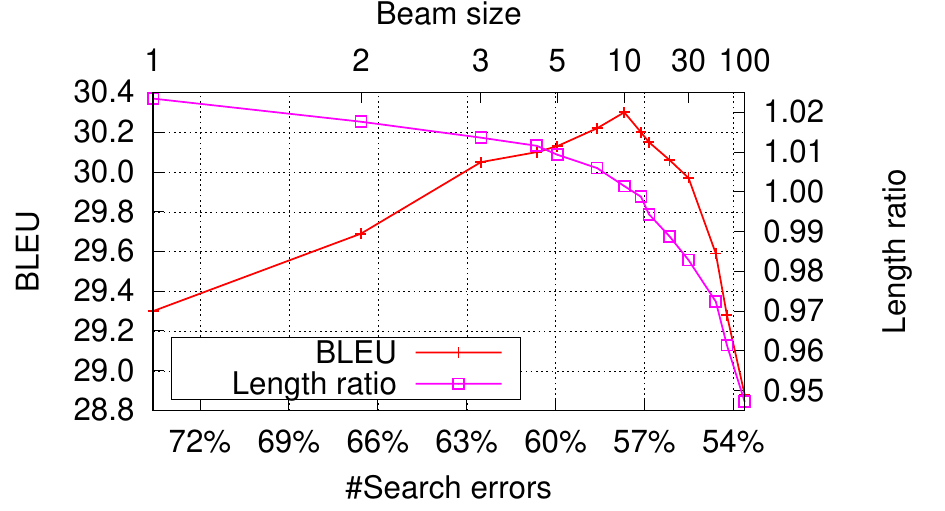}
\caption{BLEU over the percentage of search errors. Large beam sizes yield fewer search errors but the BLEU score suffers from a length ratio below 1.}
\label{fig:serrors}
\end{figure}

\begin{figure}[t!]
\centering
\small
\includegraphics[scale=0.85]{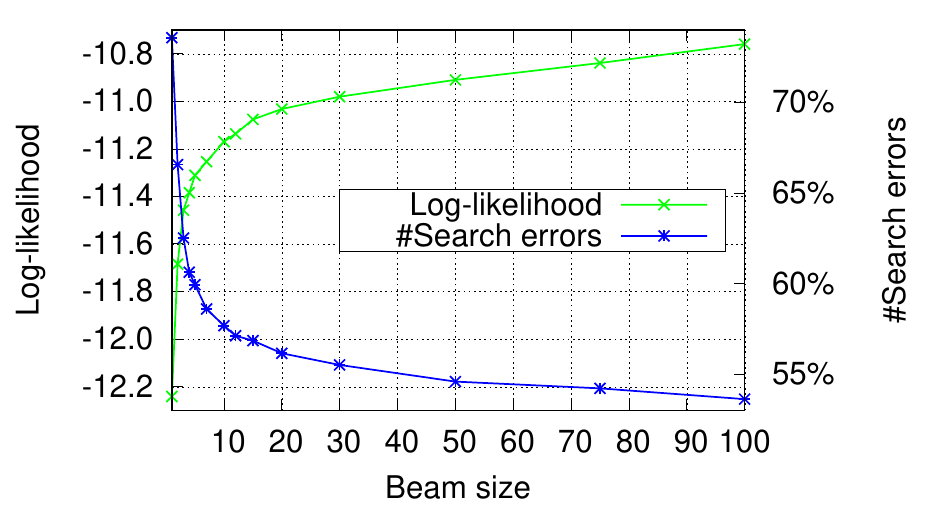}
\caption{Even large beam sizes produce a large number of search errors.}
\label{fig:beam-size}
\end{figure}

Our main result is shown in Tab.~\ref{tab:exact-search}. Greedy and beam search both achieve reasonable BLEU scores but rely on a high number of search errors\footnote{A sentence is classified as search error if the decoder does not find the global best model score.} to not be affected by a serious NMT model error: For 51.8\% of the sentences, NMT assigns the global best model score to the empty translation, i.e.\ a single $\mathtt{</s>}$ token. Fig.~\ref{fig:serrors} visualizes the relationship between BLEU and the number of search errors. Large beam sizes reduce the number of search errors, but the BLEU score drops because translations are too short. Even a large beam size of 100 produces 53.62\% search errors. Fig.~\ref{fig:beam-size} shows that beam search effectively reduces search errors with respect to greedy decoding to some degree, but is ineffective in reducing search errors even further. For example, Beam-10 yields 15.9\% fewer search errors (absolute) than greedy decoding (57.68\% vs.\ 73.58\%), but Beam-100 improves search only slightly (53.62\% search errors) despite being 10 times slower than beam-10.

\begin{figure}[t!]
\centering
\small
\includegraphics[scale=0.85]{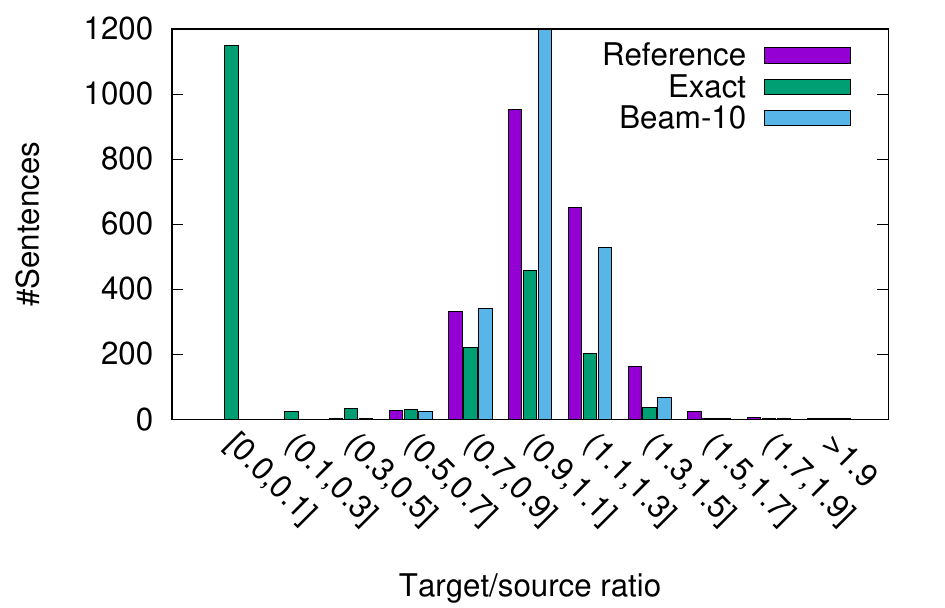}
\caption{Histogram over target/source length ratios.}
\label{fig:hist-ratio}
\end{figure}

The problem of empty translations is also visible in the histogram over length ratios (Fig.~\ref{fig:hist-ratio}). Beam search -- although still slightly too short -- roughly follows the reference distribution, but exact search has an isolated peak in $[0.0,0.1]$ from the empty translations.

\begin{table}[t!]
\centering
\small
\begin{tabular}{|l|rr|r|}\hline
\textbf{Model} & \multicolumn{2}{c|}{\textbf{Beam-10}} & \textbf{Exact} \\
& \textbf{BLEU} & \textbf{\#Search err.} & \textbf{\#Empty} \\ \hline
LSTM$^*$ & 28.6 & 58.4\% & 47.7\% \\
SliceNet$^*$ & 28.8 & 46.0\% & 41.2\% \\
Transformer-Base & 30.3 & 57.7\% & 51.8\% \\
Transformer-Big$^*$ & 31.7 & 32.1\% & 25.8\% \\
    \hline
\end{tabular}
\caption{$^*$: The recurrent LSTM, the convolutional SliceNet \citep{nmt-conv-slicenet}, and the Transformer-Big systems are strong baselines from a WMT'18 shared task submission~\citep{ucam-wmt18}.}\label{tab:wmt18-models}
\end{table}

\begin{figure}[b!]
\centering
\small
\includegraphics[scale=0.85]{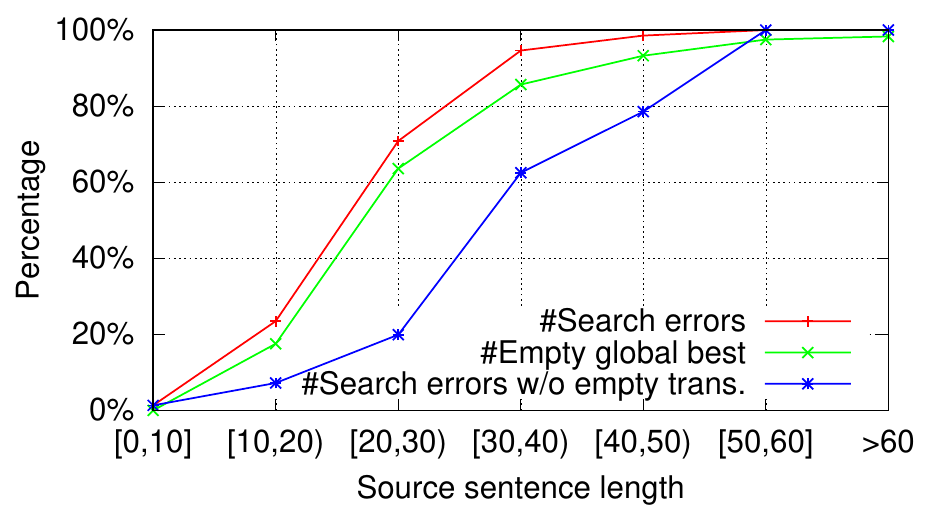}
\caption{Number of search errors under Beam-10 and empty global bests over the source sentence length.}
\label{fig:src-len}
\end{figure}

Tab.~\ref{tab:wmt18-models} demonstrates that the problems of search errors and empty translations are not specific to the Transformer base model and also occur with other architectures. Even a highly optimized Transformer Big model from our WMT18 shared task submission~\citep{ucam-wmt18} has 25.8\% empty translations.

Fig.~\ref{fig:src-len} shows that long source sentences are more affected by both beam search errors and the problem of empty translations. The global best translation is empty for almost all sentences longer than 40 tokens (green curve). Even without sentences where the model prefers the empty translation, a large amount of search errors remain (blue curve).

\begin{figure}[t!]
\centering
\small
\includegraphics[scale=0.85]{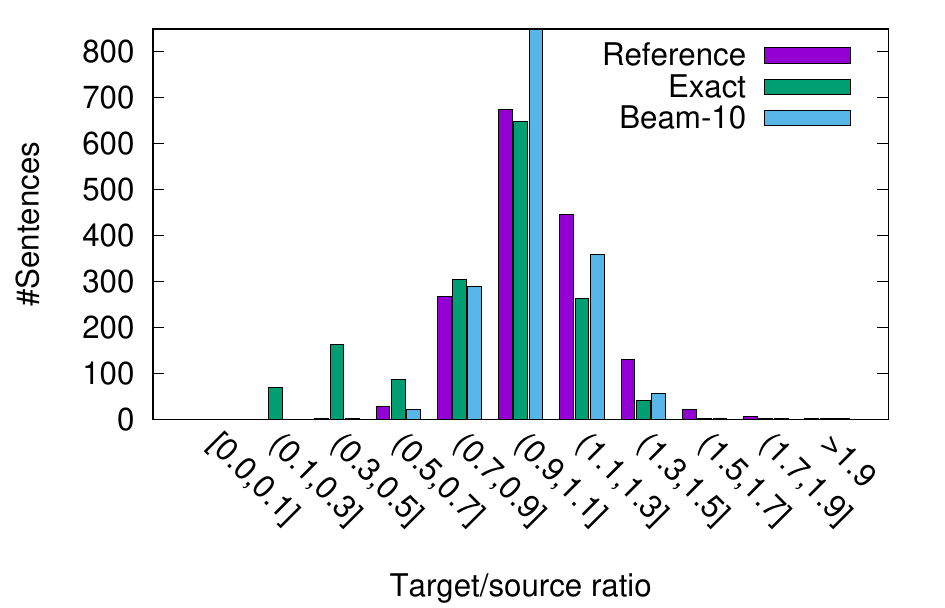}
\caption{Histogram over length ratios with minimum translation length constraint of 0.25 times the source sentence length. Experiment conducted on 73.0\% of the test set.
}
\label{fig:hist-ratio-constrained}
\end{figure}

\section{Results with Length Constraints}
\label{sec:length-constraints}

To find out more about the length deficiency we constrained exact search to certain translation lengths. Constraining search that way increases the run time as the $\gamma$-bounds are lower. Therefore, all results in this section are conducted on only a subset of the test set to keep the runtime under control.\footnote{We stopped decoding if the decoder took longer than a day for a single sentence on a single CPU. Exact search {\em without} length constraints is much faster and does not need maximum execution time limits.} We first constrained search to translations longer than 0.25 times the source sentence length and thus excluded the empty translation from the search space. Although this mitigates the problem slightly (Fig.~\ref{fig:hist-ratio-constrained}), it still results in a peak in the $(0.3,0.5]$ cluster. This suggests that the problem of empty translations is the consequence of an inherent model bias towards shorter hypotheses and cannot be fixed with a length constraint.

\begin{table}[t!]
\centering
\small
\begin{tabular}{|l|rr|}\hline
\textbf{Search} & \textbf{BLEU} & \textbf{Ratio} \\ \hline
Beam-10 & 37.0 & 1.00 \\
Exact for Beam-10 length & 37.0 & 1.00 \\
Exact for reference length & 37.9 & 1.01 \\
    \hline
\end{tabular}
\caption{Exact search under length constraints. Experiment conducted on 48.3\% of the test set.}\label{tab:length-constrained}
\end{table}

We then constrained exact search to either the length of the best Beam-10 hypothesis or the reference length. Tab.~\ref{tab:length-constrained} shows that exact search constrained to the Beam-10 hypothesis length does not improve over beam search, suggesting that any search errors between beam search score and global best score for that length are insignificant enough so as not to affect the BLEU score. The oracle experiment in which we constrained exact search to the correct reference length (last row in Tab.~\ref{tab:length-constrained}) improved the BLEU score by 0.9 points.

\begin{table}[t!]
\centering
\small
\begin{tabular}{|l|rr|rr|}\hline
\textbf{Search} & \multicolumn{2}{c|}{\textbf{W/o length norm.}} & \multicolumn{2}{c|}{\textbf{With length norm.}} \\
& \textbf{BLEU} & \textbf{Ratio} & \textbf{BLEU} & \textbf{Ratio} \\ \hline
Beam-10 & 37.0 & 1.00 & 36.3 & 1.03 \\
Beam-30 & 36.7 & 0.98 & 36.3 & 1.04 \\
Exact & 27.2 & 0.74 & 36.4 & 1.03 \\
    \hline
\end{tabular}
\caption{Length normalization fixes translation lengths, but prevents exact search from matching the BLEU score of Beam-10. Experiment conducted on 48.3\% of the test set.}\label{tab:length-norm}
\end{table}

A popular method to counter the length bias in NMT is  {\em length normalization}~\citep{sys-montreal-wmt15,nn-length-norm} which simply divides the sentence score by the sentence length. We can find the global best translations under length normalization by generalizing our exact inference scheme to {\em length dependent} lower bounds $\gamma_k$. The generalized scheme\footnote{Available in our SGNMT decoder~\citep{sgnmt1,sgnmt2} as \texttt{simplelendfs} strategy.} finds the  best model scores for each translation length $k$ in a certain range (e.g.\ zero to 1.2 times the source sentence length). The initial lower bounds are derived from the Beam-10 hypothesis $\mathbf{y}_\text{beam}$ as follows:\footnote{We add 1 to the lengths to avoid division by zero errors.}
\begin{equation}
\gamma_k=(k+1)\frac{\log P(\mathbf{y}_\text{beam}|\mathbf{x})}{|\mathbf{y}_\text{beam}|+1}.
\end{equation}
Exact search under length normalization does not suffer from the length deficiency anymore (last row in Tab.~\ref{tab:length-norm}), but it is not able to match our best BLEU score under Beam-10 search. This suggests that while length normalization biases search towards translations of roughly the correct length,  it does not fix the fundamental modelling problem.

\section{Related Work}

Other researchers have also noted that large beam sizes yield shorter translations~\citep{nmt-overview-six-challenges}. \citet{nmt-lengthbias} argue that this model error is due to the locally normalized maximum likelihood training objective in NMT that underestimates the margin between the correct translation and shorter ones if trained with regularization and finite data. 
A similar argument was made by \citet{nmt-correcting-lengthbias} who pointed out the difficulty for a locally normalized model to estimate the ``budget'' for all remaining (longer) translations. \citet{nmt-length-calibration} demonstrated that NMT models are often poorly calibrated, and that that can cause the length deficiency. \citet{conf-ana-uncertainty} argued that uncertainty caused by noisy training data may play a role. \citet{maletti} showed that the consistent best string problem for RNNs is decidable. We provide an alternative DFS algorithm that relies on the monotonic nature of model scores rather than consistency, and that often converges in practice.

To the best of our knowledge, this is the first work that reports the exact number of search errors in NMT as prior work often relied on approximations, e.g.\ via $n$-best lists~\citep{nmt-search-vs-model-errors} or constraints~\citep{sgnmt2}.

\section{Conclusion}

We have presented an exact inference scheme for NMT. Exact search  may not be practical, but it allowed us to discover deficiencies in widely used NMT models. We linked deteriorating BLEU scores of large beams with the reduction of search errors and showed that the model often prefers the empty translation -- an evidence of NMT's failure to properly model adequacy. Our investigations into length constrained exact search suggested that simple heuristics like length normalization are unlikely to remedy the problem satisfactorily.

\section*{Acknowledgments}
This work was supported by the U.K. Engineering and Physical Sciences Research Council (EPSRC) grant EP/L027623/1 and has been performed using resources provided by the Cambridge Tier-2 system operated by the University of Cambridge Research Computing Service\footnote{\url{http://www.hpc.cam.ac.uk}} funded by EPSRC Tier-2 capital grant EP/P020259/1.

\bibliography{emnlp-ijcnlp-2019}
\bibliographystyle{acl_natbib}


\end{document}